# Korean Online Hate Speech Dataset for Multilabel Classification
# - How Can Social Science Improve Dataset on Hate Speech? -

**TaeYoung Kang[1][**], Eunrang Kwon[2][*], Junbum Lee[3][*], Youngeun Nam[4][*], Junmo Song[5][*], JeongKyu Suh[6][*]**

[1]Underscore, [2,5]Department of Sociology, Yonsei University, [3]Graduate School of Data Science, Seoul National University

[4]Department of Sociology, Purdue University, [6]Department of Political Science, University of Houston

## Abstract

We suggest a multilabel Korean online hate speech dataset that covers seven categories of hate speech: (1) Race and Nationality, (2) Religion, (3) Regionalism, (4) Ageism, (5) Misogyny, (6) Sexual Minorities, and (7) Male. Our 35K dataset consists of 24K online comments with Krippendorff's Alpha label accordance of .713, 2.2K neutral sentences from Wikipedia, 1.7K additionally labeled sentences generated by the Human-in-the-Loop procedure, and rule-generated 7.1K neutral sentences. The base model with 24K initial dataset achieved the accuracy of LRAP .892, but improved to .919 after being combined with 11K additional data. Unlike the conventional binary hate and non-hate dichotomy approach, we designed a dataset considering both the cultural and linguistic context to overcome the limitations of western culture-based English texts. Thus, this paper is not only limited to presenting a *local* hate speech dataset but extends as a manual for building a more generalized hate speech dataset with diverse cultural backgrounds based on social science perspectives.

## Introduction

With the massive increase of hate speech in the online space, the ethics of data services has become an important issue for IT firms and academic institutions. Thus, we should focus on minimizing the unpleasant experience of users from malicious texts when building a chatbot or text-based model. Ahead of technological issues, there had long been a social need to build a hate speech dataset. Most of the previous works, however, included only general abusive comments under the name of *online profanity*. As these existing datasets cannot capture a variety of hate expressions occurring in online space, we should extend the categories of malicious text datasets.

The practically applicable hate speech dataset should include the following characteristics. First, it should encompass multiple categories of hate expressions. Second, even if it does not directly include swear words, it should also be able to capture the texts that contain discriminatory content. Third, the dataset not only limited to western society based English expressions is required.

The early works on hate speech identification focused on combining dictionaries and sentiment analysis (Schmidt & Wiegand, 2017). Bunde (2021) detected words indicating hate speech and Rodriguez et al. (2019) used emotion analysis to automatically unearth hate speech on Facebook data. They scored the negativity of comments through the classic sentiment analysis. On the applicational level, some papers exploited classifiers including logistic regression, random forest, and basic neural networks (Badjatiya et al., 2017; Qian et al., 2019). The largest multilingual hate speech dataset is designed by Vidgen and Derczynski (2020). It includes various languages including English, Arabic, German, Danish, etc., and provide multiclass labels including the category (gender, sexual orientation, religion, disability, etc.), level of target unit (individual vs group), and the topic (culture, economy, crime, terrorism, history, etc.)

Detecting hate speech is a challenging task due to the inherent complexity of the natural language. A single sentence can aim at multiple targets of hatred, and the cultural and local contexts are required to precisely discriminate the underlying intention of utterance. Badjatiya et al. (2017) and Kennedy III et al. (2017) labeled the dataset only with binary categories *(1=hate speech / 0=general sentences)*. Waseem and Hovy (2016) dealt with the terms *racist* and *sexist*, and Warner and Hirschberg (2012) showed more detailed categories: *anti-semitic, anti-black, anti-Asian, anti-woman, anti-muslim, anti-immigrant*, etc. Most of the previous approaches, however, lacked the reflection of local context and clarification of subtypes of hate speech based on social science theories.

[**] corresponding author
[*] equal contribution

Email : [1]minvv23@underscore.kr, {[2]eunrang_kwon, [5]cssjm}@yonsei.ac.kr, [3]beomi@snu.ac.kr, [4]nam49@purdue.edu, [6]jsuh3@cougarnet.uh.edu

Hate speech is defined as an aggressive expression against minorities and can appear in various topics such as *genders, ethnic groups, religions*, etc (Walker, 1994). Depending on the composition of the majority group in society, the target minorities of hate speech may vary. In addition, as language reflects distinct local characteristics and the process of thinking (Zhang, 2017), it is necessary to consider that each culture has its own different interpretations and characteristics in determining hate speech. For example, hate speech based on ethnicities such as race, skin color, and nationality were mainly expressed in Germany, Canada, and Denmark (Hong, 2019). Japan, on the other hand, its hate speech appears to promote discriminatory awareness against foreigners while Korea presents despise of weak minority groups such as women and the elderlies (Hong, 2019).

Based on real-world online comments, we attempt to develop a multi-label Korean hate speech. For the analysis, we conduct a literature review and design a dataset that is aware of the national and linguistic context of South Korea. By providing an exemplary approach to overcome the conventional English dataset, this paper extends as a manual for building a more generalized hate speech dataset with diverse cultural backgrounds based on social science perspectives.

## Related Work

The nature of the Internet, an anonymous space free from time and space constraints, was thought to ensure free participation for users from various social backgrounds, allowing rational discussions to take place in online spaces. However, these anonymous characteristics actually led to the emergence of malicious comments that can harm others.

*Malicious comments* are similar in type to cyber verbal violence and can be classified into abusive language, defamation, slander, sexual abuse, and rumors (안태형, 2013). The emotional damage from malicious comments can be extended to third-party targets other than those who participated in posting and comment writing. For example, racial hatreds are taboo offline, but online users tend to express their private feelings openly (Loke, 2012). Therefore, malicious comments can go beyond mere slander or abusive language and extend to hateful expressions against certain groups such as race, gender, and age.

Today, hatred is commonplace in the online space; the online communities are encouraging hatred and conflicts. The hate speech can be in the form of a *meme* to share humor in a cultural context and to reproduce stereotypes of social groups. As users exploit a strategy to turn hatred into a *play* by relieving its psychological resistance, it continues to be spread and connected with the idea that hatred is routine and justifiable (김수아 and 김세은, 2016). In this way, hate speech is formed through active interaction between the users rather than established as a specific target group itself.

*Hate speech* can be interpreted as an expression that targets the identity of minority groups that have been socially and historically discriminated and suppressed (박미숙 and 천지현, 2017). As the scope of hatred has been expanding and became commonplace in online space, hate speech could also be broadly defined as 'exclusion and social prejudice using contemptuous expressions for race, gender, region of origin, disability, etc.' (KCC, 2015). Its target groups include *region, ethnicity, race, nationality, religion, disability, gender identity and sexual orientation, age, etc.* (이성순, 2018).

Hate speech does not necessarily consist of direct and aggressive remarks. Misogynistic statements are the typical cases. Along with *hostile sexism* which expresses open hostility, *benevolent sexism* in the context of advocating and protecting femininity is also prevalent (Glick and Fiske 2001). "She married at age 30, and she did it well before it was too late" and "She must be popular because she is girly and quiet." would be an example. These benevolent hate speeches appear not only in gender but also in the area of race and nationality.\

Previous research includes finding vocabulary that refers to hate expressions (Bunde, 2021) or discovering the malice of text from online spaces including social media (Rodriguez et al, 2019; Xu, Jing, et al, 2020; Badjatiya et al., 2017; Qian et al., 2019). However, some are concerned about simply determining whether a single sentence is a hate expression or not. Chai, Yixuan et al. (2020) argues that if the goal is to determine aggression and disgust in a conversation, it is necessary to build a paired dataset. Likewise, Khatri, Chandra, et al (2018) and Qiang et al. (2019) tried to determine the hatred of the given online post by utilizing the entire history of conversations. These criticisms are valid enough as they pay attention to the characteristics of online text as bilateral or multilateral communication, but their applicability to practice remains questionable. If we exploit the paired conversation dataset, we are likely to have difficulties in applying the model to real-time conversations, as distinguishing the divergence of topics in the sequential speech data is an abstruse technological procedure itself.

Aside from this criticism, detecting hate speech is basically difficult only for the case of a single sentence. As the *memefied* hate speech is written under more complicated contexts, cultural and regional knowledge must be considered together. So far, the binary classified dataset (1=hate expression, 0=general sentence) is dominant (Badjatiya et al., 2017; Kennedy III et al, 2017), and data (Warner & Hirschberg, 2012) that provide multiple categories such as racism, misogyny, and anti-Semitism are still restricted to English-using western culture. Thus, if we want to take a more sophisticated approach than the simple detection of online profanity (Schlesinger, Ari, Kenton P. O'Hara, and Alex S. Taylor, 2018), this paper will aid in achieving such goals.

# Hate Speech Categories

Collecting comments in an online space has its clear advantage. First, it is easy to access large-scale text data. Second, offline surveys are susceptible to social desirability bias that does not want to openly express immoral ideas, while there are no such restrictions in online anonymous writings.

Online texts are often mistaken for randomly extracted data from real society, but they are not statistically independent each other unlike national level offline polls. In the case of the news portal comment section, the percentage of users writing comments is estimated to be as little as 1% to as much as 20%. (Korea Press Foundation, 2021) We can also observe strong power law patterns when analyzing the number of comments written by users, indicating that only a small number of users write most of the text in online space. Online communities are also susceptible to a filter bubble space selectively composed by certain users with clear political preferences, rather than a space where representative citizens write their idea evenly. It should also be noted that we cannot specify the demographic characteristics and identity of the author of the online comments. For example, the same user may create different accounts to repeatedly write similar posts to affect the public opinion in the community.

Hence, determining the hate speech categories and investigating their statistical distributions with fragmented online texts will easily lead to biased conclusions; We need social scientific understanding of the political and social nature of hate speech in society.

The United Nation, for example, maintained that religious and racial minorities, immigrants, children, women, and LGBTQ should be protected in the pursuit of equality and anti-discrimination. Also, National Human Rights Commission of Korea included women, sexual minorities, the disabled, patients, immigrants, black, Southeast Asian, Muslim, low-income citizens, and homeless people in its 2016[th] research (홍성수, 2016). In the 2019 Hate Discrimination National Perception Survey, more than half of the seven social groups from specific regions (74.6%), feminists (69.4%), women (68.7%), elderly (67.8%), LGBTQ (67.7%), migrants (56.0%), men (59.1%), and disabled (58.2%) were hated in both online and offline spaces (국가인권위원회, 2019). The 2016 Korean General Social Survey (KGSS), a domestic version of the General Social Survey (GSS) included foreign workers, disabled people, homosexuals from Jeolla-province, and conscientious objectors as a questionnaire of socially discriminated groups.

The most prominent Korean hate speech dataset is Moon et al. (2020). It collected 2M web portal's entertainment news comments and multi-class labeled its topic (gender-related vs general) and level of hostility (0=None, 1=Aggressive, 2=Hostile). Although the data is valuable in that it is large-scale, the boundary between *hostile* and *aggressive* is ambiguous and the comments are from articles dealing with celebrities and cultural events. 이원석 and 이현상 (2020) extends Moon et al.'s data by re-labeling 13.7K comments sampled from it and adding the five hate speech category labels – gender, politics, ageism, religion, and race. As it is basically secondary data, however, the class imbalance among these categories is observed.

Based on the review of existing studies, we set up seven categories of hate speech: (1) Race and Nationality, (2) Religion, (3) Regionalism, (4) Ageism, (5) Misogyny (Women & Family), (6) Sexual Minorities, and (7) Male.

## Race and Nationality

South Korea is well-known as an ethnically and racially homogeneous country (Alesina et al., 2003; Cederman & Girardim, 2007). Until the early 2000s, South Korea had maintained the legacy of racial and ethnic uniformed society. Thus, South Korea had not paid attention to racial issues. Moreover, they did not care which one could be regarded as offensive to racial and ethnic minority groups. For instance, it has not been difficult to find comedians who use blackface as make-up for South Korea's comedy show program and celebrities who use stereotypes toward foreigners as humor code in TV programs. However, as South Korea has gradually become a multicultural society after the early 2000s (Jang, 2010), they have taken relatively more care of racial issues than before.

Although South Korean society pays more attention to racial issues than in the past, various racial hate speeches are still detected easily (Yang, 2019). South Korea's racial hate speech is somewhat similar to Western cases in general. First, racial hate speech tends to generalize individuals' problems to specific groups' matters. In other words, people are more likely to strengthen their pre-existing negative prejudice toward specific racial and ethnic groups by using individuals' matters as their evidence. Even positive stereotypes could also be utilized as evidence of racial hate speech since it consequently can work to strengthen pre-existing stereotypes. Second, hate speech regarding race and ethnicity implies separation and exclusion based on racial, ethnic, and national in-group and out-group. Thus, racial hate speech is usually based on unconditional in-group preferences and out-group hostility. Third, hate speech to racial minorities tends to justify discrimination and prejudice against racial and ethnic minority groups by using social norms. Lastly, racial hate speech is linked to gender issues or perceived racial hierarchy in some cases. Therefore, hate speech to racial minorities has a tendency to reinforce existing stereotypes and prejudice toward specific racial, ethnic, and nationality groups by using various irrational evidence.

Unlike Western cases (e.g., Brader, Valentino, & Suhay, 2008; Hainmueller & Hopkins, 2014; 2015; Ostfeld, 2017),

racial hate speech in South Korea has a unique and distinguished aspect since it is more likely to be concentrated on ethnically and racially identical or similar immigrant groups such as Korean-Chinese (also known as *Chosunjok*), Chinese and even North Korean refugees (Ha, Cho, & Kang, 2016; Suh, 2017). This aspect seems to reflect the reality that the number of these co-ethnic immigrant groups has been continuously and consistently maintained high at an absolute and relative level. However, this aspect does not suggest that South Koreans have relatively favorable attitudes toward immigrants with different racial traits (e.g., Black, White, and Southeast Asians). South Koreans also have implicit preferences toward each racial group, leading to expressing their antagonism toward specific racial groups (Hahn et al., 2013). In addition, a recent study finds that South Koreans' antagonism toward racial minorities tends to intensify when immigrants are more likely to participate in the political arena (Suh, 2019). "Jasmine Lee (first immigrant-oriented legislator in South Korea) Hate" is the most typical and remarkable case of racial hate speech to immigrants who participated in South Korean politics. This is because supporters of the two competing political parties have shared a unified opinion in making her the target of hate speech (한겨레 21, 2015).

### Religion

South Korea has diverse religions. As a religiously diverse country, South Korea has three major religions (i.e., Buddhists, Protestants, and Catholics), but no single dominant religion exists. Also, most South Koreans do not believe in any religion. Thus, it is hard to find religious conflicts in South Korean society. However, it does not indicate that there is no religious hate speech in South Korea.

Religious hate speech in South Korea has its distinctiveness based on its unique multi-religious contexts. In most Western countries, religious hate speech usually focuses on a religious minority group in their society (e.g., Islam) because westerners tend to believe that Muslims have distinct and incompatible cultural and religious backgrounds with them (Brubaker, 2017; Lajevardi, 2020). However, South Korea's religious hate speech is mainly targeted at Protestants, one of their major religions (i.e., the religion that approximately 20% of the South Korean population believes) (강성호, 2016). Unlike the Western countries, religious minority groups such as Islam are not the main target of hate speech in South Korea. Thus, the most notable feature of religious hate speech in South Korea is that the target of religious hate speech in South Korea is not a religiously persecuted minority group. It implies that religious hate speech does not stem from unpleasantness toward religious minorities, but from the negative image of major religion which is reflected in South Koreans' stereotypes (시사저널, 2020). In other words, we can understand that the antagonism toward Protestants has emerged as an adverse reaction to their right-wing extremism and their exclusiveness toward other religions (오마이뉴스, 2019). Hate speech toward Protestants varies. It ranges from "Kae-dokkyo (Doglike Protestants)," which expresses abomination toward the entire South Korean Protestants, to hate speech for specific sect or pastors.

On the other hand, although it is not frequent than hate speech to Protestants, hate speech toward other religions such as Buddhism and Catholics is found in the online communities when negative social issues regarding specific religious groups are spotlighted. As mentioned before, hate speech to Islam is relatively not very often in South Korea. However, it easily appears if international and domestic issues related to Islam (i.e., terrorism by Muslims, inflow of Muslim refugees in South Korea) are highlighted. Especially, Islam is tended to be regarded as a negative word itself. For instance, Islam is sometimes used to disparage certain political party supporters in South Korea.

### Regionalism

In Korea's modern politics, support for political parties has been sharply divided by region; the regional characteristics have been linked with political hatred and recalled as the subject of long-standing hate speech(조기숙, 1997; Park, 2003). Also, regionalism mainly targets two specific regions – 전라도(Jeolla-province) and 경상도(Gyeongsang-province) – which are the key support area for liberal and conservative parties, respectively. Although the negative stereotypes are prevalent in various regions of Korea, they are sometimes even summoned as a serious target of hostility.

In particular, the Jeolla-province, which has been marginalized from development under the authoritarian regime and has strong support for liberal political parties, has been a major target for hate speech (박상훈, 2009). In addition, sexual crimes and slavery that occurred in the underdeveloped rural islands of the region were used to justify hatred of this region (양혜승, 2018). On the other hand, after democratization, as liberal forces have been stably established as one member of the bipartisan system, hate speech toward Gyeongsang-province, a very conservative and patriarchal region, also increased (오마이뉴스, 2021). Hate speech against these regions includes mentioning their representative foods– skate, 과메기(*gwamegi*) - or tragic events, including the Gwangju Uprising and the Daegu subway fire accident (한겨레 21, 2014).

What is particularly noteworthy about the latest regionalism hate speech in Korea is the racist sentiment which is distinct from traditional political conflicts (아시아경제, 2015). With the slowdown of economic growth of Korea and the growing influx of foreign workers, racist hatred of immigrants from Southeast Asia and China emerged in

Korean society (세계일보, 2020). Apart from hate speech against their ethnicity and nationality, hate speech against the area they reside in also started to increase. Hence, these expressions were classified as hate speech against the region.

*Daerim, Ansan, Bucheon, Incheon, Shinan... Hahaha I'm very familiar with it.*
대림, 안산, 부천, 인천, 신안…ㅎㅎ 다들 알만한 곳이구만

Most of the areas listed in the sentence above are not a problem when only considered literally, but these are the major residential area for foreign workers. Therefore, it was classified as a regionalism hate speech.

*Hey you, reporter Neoul Shim, you bastard, are you from Jeolla-province? Somehow it smells like skates;*
야 심너울 기자 이 새끼야. 너 라도 출신이냐? 어디서 홍어 냄새가 나더라.

In addition, as in the above sentence, the case in which other social groups, such as occupational groups, were associated with local issues and were slandered was also determined as local hate speech.

### Ageism

Hate speech against a specific generation can be expressed as ageism. Ageism is defined as "prejudice by one age group toward other age groups" (Butler, 1969). This concept indicates an idea, attitude, or behavior that discriminates based on age and includes both *Ephebiphobia* and *Gerontophobia*. If social norms against expressing Gerontophobia are strong, discrimination against older people may not be expressed publicly (Crandall, Eshleman, & O'Brien, 2002). On the other hand, Gerontophobia can intensify in the online space because most users are young, and anonymity weakens the real-world norms. Age discrimination in online spaces such as social media is common in derogatory language directed at groups or individuals (Kroon et al., 2019). Social media shapes an ageist public opinion (Qazi & Shah, 2018).

Ageism about older people is mainly due to their physical status, the difference in political beliefs compared to other generations, and the burden of social support for the elderly. In the case of physical characteristics, older people are assumed to be in an unproductive and worthless position, so they are likely to become the object of hate that gives fear to members of society (Butler, 1989). Ageism in young adults is an unconscious defensive strategy to relieve death anxiety (Bodner, 2009). A study about hate speech in online space shows that there were many cases of slandering the older people as a group with a different opinion and political beliefs than younger people and not able to communicate. This study also found slander against the older people because of the appearance of ignoring social order and norms (Shin & Choi, 2020). In addition, hate speech for older people occurs because they deprive the younger generation of opportunities and becomes a social burden (Braceland, 1972).

The severity of ageism can vary according to national and cultural contexts. Ageism may intensify in countries experiencing a rapid development and aging because Eastern countries such as Korea, which achieved an abrupt industrialization and economic growth, differences in values between generations can widen. Previous literature suggests Eastern cultures are less ageist than Western cultures because they emphasize high esteem for older people (Nelson, 2009). However, recent studies tell that it is important to consider the normative context and personal attitudes rather than supporting such stereotypes (Vauclair et al., 2017). This weakened value of respect and North and Fiske also shows the rapid increase in the older population can intensify the conflict between the younger and older generation.

Due to the nature of online communication, there seems to be virtually no way to find out who is the elder, but language expressions that disparage others like 꼰대(ubervisor) and 틀딱(boomers) begun to appear in Korea. These derogations are usually attached to both *586 generation*, the South Korean counterparts of `68 generation in Europe and *Industrial generation*, the elderlies who miss the authoritarian government in the early days of Korean society.

Carl Manheim, who presented the first sociological research framework for generations, classified the concept of generation into *generation location*, *generation as actuality*, and *generation unit*. (Mannheim, 1970) Based on this conceptualization, diverse social scientists in Korea have been focusing on the praise, criticism, sympathy, and conflicts regarding the *generation as actuality*. (전상진, 2019; 이철승, 2019; 김창환 and 김태호, 2020; 신진욱, 2020)

Therefore, considering the characteristics of each country, it is important to consider hate speech against age not only in Western culture but also in Eastern cultures, which are conventionally known to value respect for older people.

### Misogyny (Women & Family)

It is imperative to develop a dataset implementing sociological perspectives when identifying gender discrimination online because gender refers to a social category represented with femininity and masculinity (not based on a biological difference) (West and Zimmerman 1987). Since gender is a social concept, what it means to be feminine and masculine has changed over time. Historically, women were deemed feminine by being subservient, caring, and motherly. on the other hand, in modern society women have also partaken in a domain that used to be considered traditionally masculine– e.g. employment. Because the traditional and modern values of femininity are contradictory, women are often criticized for not fitting into either type of femininity (Hardy 2015).

In Korea, misogyny–discrimination and oppression of women– takes place because of precarity between femininity and masculinity and is often subtle and invisible (김영희 2018). For example, there have been moves to return to an androcentric and patriarchy society on online communities like *Kimchi-nyeo* on Facebook, referring to Korean women [nyeo] in a derogatory manner, and *Il-be* (김수아 and 김세은 2016). For another example, Korean male online users frame women framed as abnormal and irrational, while portraying men as rational and innocent victims who need to sacrifice for the family, instead of acknowledging gender inequality in Korean society. In other words, although there is social awareness against hostile sexism–such as gender violence, objectification of women and sexual assault, subtle forms of sexism are still rampant in Korea, especially online. Because diverse forms of hatred against women are prevalent in Korean online communities (정인경 2016), it is critical to develop data sets that accounts for not only the nuance in the language but also its social context.

Sociological analysis in developing and analyzing online discrimination regarding family is crucial because family is a social organization but also an important foundation of society that has been historically formed by heterosexual marriage and blood. At the same time, family patterns have been changing across the world due to aging, declining fertility rates, and other societal factors.

In European and North American countries, a family takes diverse forms including cohabitating families, same-sex marriage, re-marriage, divorce, single-household, and more. In Korea, there also has been an effort to introduce diversity in family patterns by challenging heteronormative marriage (김순남 2016). Abolition of the adultery law, which used to criminalize extramarital affairs, in 2015 reflects such change in Korea. However, discrimination against non-normative families in Korea is overt and explicit (차유정 and 강선경 2020) because of its strong emphasis on heterosexual marriage, pregnancy, and childbirth. Unfortunately, families that diverge from such type of family are labeled as "abnormal" or "vulnerable" social groups. In this context, Korea is different from European and North American contexts countries are more inclusive of diverse family patterns.

**Sexual Minorities**

Sexuality refers not only to sexual behavior and attitude but also to sexual orientation. Although 29 countries legalized same-sex marriage, only one of them (Taiwan) is in Asia. Korea is certainly not a country that provides any legal or social protection for sexual minorities. In Korea, social support for sexual minorities – lesbian, gay, bisexual, transgender, and queer – has grown since the democratization movement in the 1980s and the first "Queer Culture Parade" in 2000 which has occurred annually ever since. However, discrimination against same-sex relations and sexual minorities is still explicit in Korea. Those who discriminate justify their actions by highlighting the unique context of Korea, arguing that Korea is different from other [Western] countries that (정애경 and 윤은희 2020).

In sum, because Koreans rationalize discrimination against sexual minorities based on the "unique" context of Korea, it is also important to implement the context of Korea, at least different from Western contexts, when developing the dataset on online hate comments against sexual minorities.

**Male**

Hostile remarks against the male are one of the most rapidly emerging types of aggressive online comments in South Korea since the late 2010s (KBS, 2016). The online space of south Korea has been male-dominated space where misogynistic speech was prevalent, just like the other countries. However, the new generation of women embarked on a massive strike back (김민정, 2020). The younger generation of women who are familiar with the Internet just as their male counterparts built their own online community in cyberspace and expressed their grievances through collective action (송준모 and 강정한, 2018). The fierce cyber warfare against male users who still insist on sexist values with misogynistic attitudes has radicalized them. Meanwhile, the alt-right ideology emerged from young men who became obsessed with 'war' against feminism. As a result, counterattacks by women who were enthusiastic about cyber warfare also intensified, and various male-hating expressions emerged (박대아, 2018).

It is controversial whether using male-hating expressions can be considered as *hate speech*. The key argument is that the hostile remarks against males cannot be conceptually equivalent to misogyny due to the asymmetry of the power in gender relationships. But if we adopt more comprehensive definition of hate speech as "stigmatizing a specific group with an undesirable attribute and legitimately expressing hostility" (Howard, 2019), expressing hatred about men in online space can be classified as hate speech, independent of the power relationship among the genders in the real world.

For instance, the following sentence that stigmatizes men as cartels and inciting violence were classified as hate speech against the males.

*Cockfucker, is it fun? testicles's solidarity'? You guy's head must be broken.*
한남충들아, 알탕연대 재밌냐? 니들은 대가리를 깨야 해.

However, if an expression that stems from this 'cyber anti-sexist warfare' was just used as a general slang instead of precisely targeting males, they were classified as simple malicious comments instead of hate speech. The slang 재기(jaegi) in the sentence below originally refers to an anti-feminist activist who committed suicide. Due to the social interest in his suicide, online communities began to use it as a meaning of *suicide of biological male groups* several years ago (쿠키뉴스, 2018). But now, it is just used as a slang to ridicule the concepts such as 'death' and 'fall'. As sociologist Pierre Bourdieu pointed out earlier, this aspect of linguistic activity in the online community is not so unusual because the *logic of practice* (Bourdieu, 1990) used by a particular group is more of a custom than the preservation of its own meaning. Therefore, we only judge the following sentences as malicious comments, instead of hate speech.

*In the first and second semesters of high school, my grades were jaegi and I couldn't recover.*
고등학교 첫 해 성적이 재기해버려서 회복이 안 되네.

## Data and Experiment

### Data Collection

A total of 24K online comments were collected from the major web portals' news section (*Naver* and *Daum*) and online community websites (*DCInside*, *Ilbe*, *Womad*, and *Today Humor*) in South Korea.[1]

In the case of news comments, 54 major newspaper companies' 20K articles written between January 2019 and June 2020 that include at least one of the 48 hate speech keywords in their titles were collected. (See *Appendices* for the details) We collected these articles through *BigKinds*, the largest official Korean News Database run by Korea Press Foundation. Since all the major news articles are also uploaded in the news aggregator portals, we matched 13K articles with their web portal uploaded versions in *Naver* and *Daum*, the two largest web portals in South Korea.

*DCInside* is the largest online community in South Korea. It has *galleries* that serve the exact function as the *subreddits* of the renowned online community service *Reddit*. Although the atmosphere and the culture differ hugely in each gallery, the major proportion of them are male-dominated and exposed to provocative and aggressive remarks. *Ilbe* is the third most popular online community in Korea that is mainly used by right-wing extremists. (김학준, 2014) Its users' main targets of hate speech are regionalism, feminism, and sexual minorities. *Womad* is a TERF (Trans-Exclusionary radical feminist) community, which adheres to an extreme position on gay and Male-to-Female transgenders. It is socially well known for jargon to ridicule males. *Today Humor* is a pro-liberals community unlike the other websites discussed above, but often presents benevolent sexism when its members discuss gender issues.

These 24K comments were divided into a total of 10 groups, and then three annotators were allocated to label the comments. The total number of annotators was 13, including the authors of this paper, all of which consisted of researchers with a master's degree or higher in social science. The final label is determined by a majority vote among three labels, and the consistency between labels was quantitatively evaluated through Krippendorff's Alpha.

With seven hate speech categories, one general profanity category, extra hate speech, and the last ordinary harmless text, the label accordance among these 10 multi-label data achieved Krippendorff's Alpha of .713. Considering the fundamental problem of the ambiguous boundary between general profanity and extra hate speech, the accordance can be evaluated as quite satisfactory.

### Models

We trained three kinds of multilabel classifiers: KcBERT-base, KcBERT-large (Lee, 2020), and KcELECTRA-base (Lee, 2021). KcBERT and KcELECTRA are pretrained masked language models which are trained on the Korean online news comments and outperform on benchmark datasets from noisy user-generated texts, showing the state of the art on NSMC (Naver sentiment movie corpus) dataset[2] and Korean Hate Speech Dataset (Moon et al., 2020). Since the domain of the corpus to train KcBERT and KcELECTRA is similar to our new dataset, we expect that these pretrained models should achieve the most successful results. We fine-tuned the pretrained language models using the Huggingface(Wolf et al., 2019) library, appending an additional linear layer after the [CLS] token.

For all models, a BERT WordPiece tokenizer was adopted, in which KcBERT-base and KcBERT-large have 30000 and KcELECTRA-base has 50135. For all models, we set the maximum length to 160 to cover all comments in size. We fine-tune each model up to 20 epochs and choose the checkpoint which shows the highest LRAP score. For the base size models, we adopt 64 as the batch size, and for the large size model, we adopt 16 as the batch size. We set other options the same with learning rate with 1e-5 with Adam optimizer using weight decay with 0.0001.

When classifying texts using deep learning models, models tend to classify the texts using the existence of specific keywords or tokens, not fully understanding the underlying contextual harmfulness. For instance, whatever the context, if the keywords *Gay*, *Chinese, Filipinos*, and *Skate*[3] were used in the sentence, then the initial classifier categorized the sentence as harmful. Hence, to prevent the overestimation of hate speech in the real-world data, we added 2.2K neutral sentences from Wikipedia dealing with hate speech issues, rule-generated 7.1K neutral sentences, and 1.7K additionally labeled sentences generated by HITL(Human-in-the-Loop) procedure.[4]

---

[1] https://github.com/smilegate-ai/korean_unsmile_dataset
[2] https://github.com/e9t/nsmc

[3] Skate is cooked in a fairly unique way in *Jeolla* province, so it is used as slang to depict the people who are from socially discriminated *Jeolla*.

**Metrics**

We adopt the label ranking average precision (LRAP) as a primary metric for our model when learning the data with 3:1 train-test ratio. LRAP is the average over each ground truth label assigned to each sample of the ratio of true vs. total labels with lower score. This metric is used in the multilabel ranking problem, where the goal is to give a better rank to the labels associated with each sample. The obtained score is always strictly greater than 0 and the best value is 1.

**Results and Examples**

| LRAP Score | basic data | HITL added |
| --- | --- | --- |
| KcBERT-base | .886 | .914 |
| **KcBERT-large** | **.892** | **.919** |
| KcELECTRA-large | .884 | .912 |

Table 1 : LRAP Score of Benchmark Experiments

Table 1 depicts LRAP scores of the three benchmark classifiers. Compared to the three classifiers, KcBERT-large model achieved the highest LRAP score, suggesting the larger size of the model leads to better performance. Nevertheless, the error rate between the classifiers is less effective than the performance gap between the 24K basic data and the 34.2K HITL based dataset meaning that the human curated negative samples are more helpful to the model to understand the underlying hateful context.

| Example | basic model | HITL model |
| --- | --- | --- |
| That Guy is Chinese<br>저 사람 중국인이네 | .867 (race) | .196 (race) |
| Are you feminist?<br>너 페미니스트니? | .028 (women) | .006 (women) |
| Same-sex marriage is controversial.<br>동성혼은 논쟁적이지 | .347 (LGBT) | .008 (LGBT) |
| You're going to kill all Muslims?<br>무슬림을 다 죽인다고? | .835 (religion) | .761 (religion) |

Table 2 : Model Comparison Examples

The performance of adding neutral documents and human-in-the-loop data to the model is stated in Table 2. When the model only learned initial data, it misclassified several sentences that mention controversial social groups as hate speech. Even for the case in which both models show equivalent decisions based on the .5 probability threshold, HITL model shows better performance compared to the basic model.

| Example | prob>.5 |
| --- | --- |
| Hey girls. Go back home and just care your child.<br>여자는 집에서 애나 봐라 | .858 (women) |
| Korean codgers, just die.<br>상폐 한남들 다 재기하라고 | .551 (ageism)<br>.877 (male) |
| Wow, a pervert festival in the city center.<br>도심에서 변태성욕 축제라니 | .810 (LGBT) |
| Don't you fat feminazis also hate Filipino beggars?<br>쿵쾅이들도 필리핀 그지는 싫지? | .740 (women)<br>.706 (race) |
| Christians are just like Chinese.<br>개신교인이나 중국인이나 거기서 거기 | .576 (religion)<br>.727 (race) |

Table 3 : Example Sentences

The example sentences are summarized in Table 3. The phrases *pervert festival in the city center* and *Christians are just like Chinese* are high-context comments as it indicates the negative perception on queer culture festival, and derogatory comparison of two controversial social groups, respectively. *상폐(delisting) 한남들(Korean males)*, a slang to ridicule men who aren't young any more, which does not directly mention the concept of aging is still captured as a malicious ageism comments. Overall, strength of multilabel classification that does not miss multiple categories is observed.

**Conclusion**

The practically applicable hate speech dataset should encompass multiple categories of hate expressions, include sentences that implicitly express socially undesirable ideas, and should not be limited to a western society based English expressions.

This paper conducts a theoretical discussion on the concept and categories of hate speech based on social science theories and designs a dataset that is aware of the national and linguistic context of South Korea. Based on the review of existing studies, we set up seven categories of hate speech: (1) Race and Nationality, (2) Religion, (3) Regionalism, (4) Ageism, (5) Misogyny, (6) Sexual Minorities, and (7) Male.

We suggested a multilabel Korean online hate speech dataset that covers various terms surrounding social minorities. Our 35K dataset consists of 24K online comments with Krippendorff's Alpha label accordance of .713, 2.2K neutral sentences from Wikipedia, 1.7K additionally labeled sentences generated by the Human-in-the-Loop procedure, and rule-generated 7.1K neutral sentences. The base model with 24K initial dataset achieved the accuracy of LRAP .892, but improved to .919 after adding 11K additional data.

---
[4] https://github.com/sgunderscore/hatescore-korean-hate-speech

By providing an exemplary approach to overcome the limitation of the conventional English dataset, this paper could be used as a manual for building a more generalized hate speech dataset with diverse cultural backgrounds based on social science perspectives.

## Acknowledgements

This study was financially supported by *Smilegate AI center* of *Smilegate Entertainment*. We also thank following annotators for their participation and advice on this research project. Without these experts in social science, we wouldn't be able to build a high-quality hate speech dataset.

Bomi Lee (Department of Political Science, Sogang University, bomipeace@gmail.com)
Hakjoon Kim (Department of Sociology, Seoul National University, sakkamot@snu.ac.kr)
Hyeyoon Kwon (Department of Anthropology, Seoul National University, hyeyoon0329@snu.ac.kr)
Hyungjun Park (Department of Psychology, National University of Singapore, e0437677@u.nus.edu)
Jiyoung Hwang (Department of Sociology, Sogang University, jyhwang@sogang.ac.kr)
Keyeun Lee (Department of Communication, Seoul National University, kieunp@snu.ac.kr)
Soyeon Ji (Department of Sociology, Sogang University, syistliebt@naver.com)
Soomin Hong (Department of Political Science, Yonsei University, hsuuum25@gmail.com)
Sungil Lee (Department of Sociology, Sogang University, leesungal@naver.com)

## Appendices

### List of Newspaper Companies

경향신문, 국민일보, 내일신문, 동아일보, 문화일보, 서울신문, 세계일보, 조선일보, 중앙일보, 한겨레, 한국일보, 매일경제, 머니투데이, 서울경제, 아시아경제, 아주경제, 파이낸셜뉴스, 한국경제, 헤럴드경제, 강원도민일보, 강원일보, 경기일보, 경남도민일보, 경남신문, 경상일보, 경인일보, 광주매일신문, 광주일보, 국제신문, 대구일보, 대전일보, 매일신문, 무등일보, 부산일보, 영남일보, 울산매일, 전남일보, 전북도민일보, 전북일보, 제민일보, 중도일보, 중부매일, 중부일보, 충북일보, 충청일보, 충청투데이, 한라일보, KBS, MBC, OBS, SBS, YTN, 디지털타임스, 전자신문

### List of Search Queries - Hate Speech Keywords

틀딱, 급식충, 잼민이, 틀니딱딱, 노인 혐오, 아동 혐오, 퀴어, 퀴어퍼레이드, 동성애, 게이, 트랜스젠더, 레즈비언, LGBT, 성소수자, 차별금지, 미투운동, N번방, 성폭행, 성매매, 이남자, 20대 남자, 한남충, 예멘 난민, 다문화, 조선족, 탈북자, 반일, 반중, 이자스민, 이민자, 무슬림, 이주민, 이민, 페미니즘, 페미니스트, 젠더, 성차별, 여성 인권, 미혼모, 지역주의, 지역 차별, 지역 혐오, 대형교회, 목사, 무슬림, 이슬람, 개신교, 개독교

### List of Neutral Wikipedia Document Titles

5·18 광주 민주화 운동, LGBT, 가부장제, 가톨릭, 갈비 구이, 강간, 강원도, 개신교, 게이, 결혼, 경상남도, 경상도, 경상북도, 계급 차별, 과메기, 광주시, 광주, 교토, 교황, 교회, 구미, 구이, 국가인권위원회법, 국민국가, 국민의힘, 국제 여성의 날, 군대, 군형법상 추행죄, 그리스도, 급식, 급진여성주의, 기도, 기독교, 기독교 여성주의, 기독교의 교파, 기안 84, 김대중, 김영삼, 김일성, 김정은, 김정일, 꾸란, 나라별 법률에 따른 성소수자의 권리, 나주, 난민, 남북관계, 남성우월주의, 남성주의, 남성학, 남성혐오, 노무현, 노태우, 노회찬, 뉴욕, 니카라과, 다문화가정, 다문화교육, 다문화주의, 다신론, 담양, 대구, 대구 지하철 화재 참사, 대구광역시, 대한민국, 대한민국 국군, 대한민국 국군의 성적 지향 및 성정체성, 대한민국 국군의 트랜스젠더, 대한민국 내 인종차별, 대한민국 여성가족부 관련 논란, 대한민국 지방자치법, 대한민국 차별금지법, 대한민국-중화인민공화국 관계, 대한민국의 병역 제도, 대한민국의 성 불평등, 대한민국의 지역주의, 대한민국의 트랜스젠더, 더불어민주당, 덴노, 도시, 도쿄, 동남 방언, 동남아, 동남아시아, 동북 방언, 동성결혼, 동성애, 동성애 혐오, 동성애와 종교, 디시위키, 러시아, 러시아인, 레즈비언, 류시민, 리비아, 마라탕, 말레이시아, 망명, 메갈리아, 메리 울스턴크래프트, 명성교회, 목사, 무슬림, 무신론, 무함마드, 문재인, 물리치료, 미국, 미국인, 미혼모, 민족주의, 바른미래당쌍용자동차, 바티칸, 박근혜, 박정희, 반일, 반일 감정, 반중, 반한 감정, 백인, 베네수엘라, 베트남, 베트남어, 병실, 병역, 부산, 부천, 북한, 불가지론, 불교, 비혼, 사우디아라비아, 사제, 사회주의적 여성주의, 새끼줄, 서귀포, 서남 방언, 서울, 성 차별, 성경, 성매매의 여성주의적 관점, 성소수자, 성역할, 성적 대상화, 성적 지향 및 성정체성과 병역, 성적 지향 차별, 성전환 치료, 성차별, 성추행, 성폭행, 섹스, 소득, 소망교회, 쇼비니즘, 수녀, 수니파, 수도원, 순창, 스님, 스시, 스위스, 시골, 시리아, 시아파, 시진핑, 신, 신나치주의, 신부, 신안, 신앙, 신학, 심상정, 싯다르타, 싱가포르, 싱가폴, 아라비아, 아베, 아시아, 아파르트헤이트, 아프가니스탄, 안철수, 알라, 야훼, 양성애, 양성애 혐오, 양육, 에이즈, 여성 할례, 여성가족부, 여성공포증, 여성의 여러 권리에 대한 옹호, 여성주의, 여성주의 미술, 여성주의의 운동과 이념, 여성학, 여성혐오, 연령 차별, 영남 지방, 영어, 예멘, 예수, 오사카, 외모 차별, 우생학, 우울증, 울산, 워마드, 워싱턴, 원희룡, 유대인, 유리천장 현상, 유승민, 유시민, 유신론, 유학, 윤석열, 이낙연, 이라크, 이명박, 이민, 이스라엘, 이슬람, 이승만, 이신론, 이자스민, 이집트, 인간 혐오, 인권, 인도, 인도네시아, 인도인, 인종, 인종국민주의, 인종본질주의, 인종의 용광로, 인종차별, 인천, 인플레, 일간베스트저장소, 일베, 일본, 일본어, 일본인, 임관, 임신, 임플란트, 자유주의적 여성주의, 자유한국당, 장교, 장애인, 장애인 차별, 전두환, 전라남도,

전라도, 전라북도, 전역, 전주, 정신병, 정의당, 정치적 올바름, 제노포비아, 제주, 젠더, 젠더리즘, 젠더퀴어, 젠더폭력, 젠더학, 조국, 조선족, 조현병, 종교, 종교 차별, 종교학, 주디스 버틀러, 주디스버틀러, 중국, 중국어, 중국인, 중동, 증오범죄, 증오언설, 지방, 지방자치, 지역, 지역주의, 직업 차별, 차르, 차별금지법, 참깨, 천동설, 천안, 천황, 초밥, 초원복집 사건, 추미애, 출산, 충청남도, 충청도, 충청북도, 친일, 친중, 카레, 카톨릭, 캄보디아, 퀴어, 퀴어 이론, 퀴어퍼레이드, 퀴어학, 크리스찬, 크리스천, 탈북자, 태국, 토박이, 트랜스젠더, 트랜스젠더와 병역, 트랜스포비아, 특별시, 틀니, 팔레스타인, 페미니스트, 페미니즘, 편견, 포항, 하나님, 하느님, 학력, 한국어의 방언, 한국의 순혈주의, 한국의 여성주의, 한국의 지방 구분, 한국의 페미니즘, 한일관계, 해운대, 혐오발언, 혐오표현, 호남 지방, 홍어, 홍준표, 화이트 내셔널리즘, 환자, 황교안, 황인, 회, 흑인, 히스패닉, 히틀러